\documentclass{article}

\PassOptionsToPackage{numbers,compress}{natbib}

\usepackage[main, final]{neurips_2026}

\makeatletter
\renewcommand{\@noticestring}{AI Agents for Discovery in the Wild (AID-Wild), Workshop at ACM CAIS 2026.}
\makeatother

\usepackage[utf8]{inputenc}
\usepackage[T1]{fontenc}
\usepackage{hyperref}
\usepackage{url}
\usepackage{array}
\usepackage{booktabs}
\usepackage{placeins}
\usepackage{amsfonts}
\usepackage{amssymb}
\usepackage{amsmath}
\usepackage{pifont}
\usepackage{enumitem}
\usepackage{nicefrac}
\usepackage{microtype}
\usepackage[dvipsnames]{xcolor}
\usepackage{graphicx}
\usepackage[most]{tcolorbox}

\colorlet{my-red}{BrickRed!90!Sepia}
\definecolor{boxaccent}{HTML}{4E3629}
\tcbset{
  takeawaybox/.style={
    width=\linewidth,
    boxrule=0.8pt,
    top=6pt,
    bottom=2pt,
    left=3pt,
    right=3pt,
    before skip=1pt,
    after skip=1pt,
    colback=boxaccent!3!white,
    colframe=boxaccent,
    colbacktitle=boxaccent,
    enhanced,
    breakable,
    attach boxed title to top left={yshift=-0.1in,xshift=0.15in},
    boxed title style={boxrule=0pt,colframe=white,},
  }
}
\newtcolorbox{takeawaybox}[2][]{takeawaybox,title={#2},#1}

\title{Declarative Data Services: Structured Agentic Discovery for Composing Data Systems}

\author{%
  Shanshan Ye\thanks{Equal contribution.} \\
  Northeastern University \\
  \texttt{ye.sha@northeastern.edu}
  \And
  Duo Lu\footnotemark[1] \\
  Brown University \\
  \texttt{duo\_lu@brown.edu}
}

\begin{document}

\maketitle

\begin{abstract}
Agentic discovery has shown that LLM-driven search can find novel algorithms, designs, and code under benchmark conditions. 
Translating the paradigm to multi-system data backends surfaces a harder problem: the search space is heterogeneous, 
the verifier is whether a deployed stack actually runs, and composition knowledge is unevenly captured in pretraining. 
\emph{Unbounded agentic discovery}, a coding agent iterating on failure-log feedback, fails to converge consistently 
on a working stack even when iteration and explicit composition knowledge are added.
We propose Declarative Data Services (DDS), an architecture for \emph{structured agentic discovery} of data-system compositions 
from declarative user intent. The framework owns four typed contracts at successive layers
(intent, operator DAG, per-system skills, runtime attribution) that decompose the global search into bounded sub-searches; 
sub-agents search each typed space, while the framework provides the channels by which knowledge flows forward as 
inline skill citations and errors route backward as typed signals. As a proof of life on a trading-backend workload, 
DDS converges where unbounded discovery does not; runtime failures become skill patches that the next deployment cites inline. 
We position this as an early prototype reporting lessons from real-world data-system composition.
\end{abstract}

\section{Introduction}
\label{sec:intro}

Agentic discovery has made a fast transition from research to product. AlphaEvolve~\citep{novikov2025alphaevolve} discovered novel algorithms by LLM-driven evolutionary search; EvoX~\citep{liu2026evox} and AdaEvolve~\citep{cemri2026adaevolve} meta-evolved the search strategies themselves for systems-discovery problems; GEPA~\citep{agrawal2026gepa} optimized declarative LM-module graphs by reflective natural-language attribution; Glia~\citep{hamadanian2025glia} produced expert-level distributed-systems designs by multi-agent reasoning; agentic coding tools such as Claude Code~\citep{anthropic2025claudecode} moved single-repository code generation into product-grade workflows. The common shape across these systems is \emph{structured agentic search}: an LLM-driven agent explores a typed solution space, a verifier evaluates each candidate, and the loop refines under structured feedback.

Translating this paradigm to multi-system data backends, the kind of stack a small team or individual without a dedicated data-engineering function 
(i.e., a solo trader, a small product team, a research group) would compose to run a real workload, surfaces a different set of problems. 
The scope is end-to-end: composing across queues, OLAP and OLTP stores, caches, search indices, and the connectors and configurations between them, 
then deploying that composition as a working stack and evolving it as the workload moves. The search space is heterogeneous; 
the verifier is whether the deployed stack actually runs and meets declared SLOs, not a benchmark answer key; ground truth is partial; 
and composition knowledge that distinguishes a working stack from a broken one is unevenly captured in pretraining and changes with every release cycle. 
Adjacent automation searches narrower spaces: infrastructure-as-code renders a chosen architecture, 
self-driving DBs tune inside one product, and modern data-stack tools declare within one pipeline stage. 
The search above these (which topology, which products, which configurations actually work together) is the gap DDS targets. 
The question is not whether agents can codegen, but how the structure around them should be designed: \emph{how should typed abstractions and coding agents split responsibility to discover, deploy, and evolve data backends from user intent?}

\begin{figure}[t]
\centering
\includegraphics[width=0.60\linewidth]{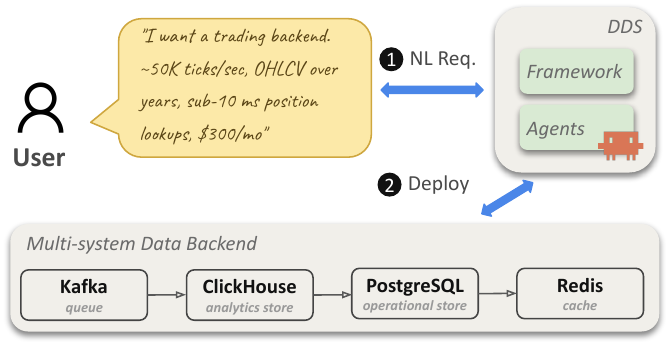}
\caption{End-to-end view of DDS. The user states intent in natural language with concrete constraints; DDS sits between intent and deployment as a structured-discovery framework; the output is a multi-system backend whose components specialize across the workload.}
\label{fig:overview}
\end{figure}

Figure~\ref{fig:overview} sketches our answer. The framework owns typed contracts at four layers (L1--L4); sub-agents search each layer's typed sub-space. Each contract is \emph{validatable}, \emph{citable}, and \emph{editable}, properties that let knowledge pass forward through the layers as the search progresses and let runtime errors route backward to the layer owning the violated decision; \S\ref{sec:framework} develops the architecture.

We make three contributions. (i)~An architecture for structured agentic discovery of multi-system data backends: a framework/agent split with four typed contracts at L1--L4, governed by two ownership rules (the framework owns each layer's contract and validation; sub-agents own the bounded search inside that contract; \S\ref{sec:framework}). (ii)~Two architectural ideas that make discovery structured rather than unbounded: typed attribution at L4 routes every runtime signal to the layer owning the violated decision, and agent skills are the persistent memory where composition knowledge accumulates (\S\ref{subsec:skills}, \S\ref{subsec:attribution-and-learning}). (iii)~A trading-backend case study (\S\ref{sec:case-study}, proof of life), with skill-content ablation and learning-loop demonstration; a second domain (chat, Appendix~\ref{app:chat}) describes operator-algebra extensibility.

\section{Motivation and Scope}
\label{sec:motivation}

\paragraph{A running scenario.}
We anchor the discussion to the trader workload of Fig.~\ref{fig:overview} (full intent in \S\ref{sec:case-study}): a real-time analytics backend combining high-throughput streaming ingest, multi-year time-series history, low-latency operational lookups, and a small budget. The user turns to a coding agent such as Claude Code; on a single repository the agent succeeds, on this multi-system stack it fails to converge consistently (empirical evidence in \S\ref{subsec:headline} and Appendix~\ref{app:failure-analysis}). This is one instance of a broader class: agentic discovery for real-world data-system composition, where the search space is heterogeneous and the verifier is whether a deployed stack actually runs.

\paragraph{Adjacent work covers slices, not structured discovery.}
Adjacent directions search narrower spaces than DDS targets. Polystores~\citep{duggan2015bigdawg} search at query time across an already-composed set of stores, not the composition itself. Self-driving DBs~\citep{vanaken2017ottertune} search the configuration space within one product, not across products. Modern data-stack tools (dbt~\citep{dbt}, Airbyte~\citep{airbyte}, Fivetran~\citep{fivetran}) offer declarative surfaces within one pipeline stage but do not compose cross-stage dataflow. Vendor data platforms (Snowflake~\citep{dageville2016snowflake}, Databricks Lakehouse~\citep{armbrust2021lakehouse}) lock the user into one vendor's product mix and are not neutral over the 400+ database systems in production~\citep{dbengines2026}. Agent-first data-system redesign~\citep{liu2026agentfirst} redesigns databases \emph{for} agents; the complementary direction, agent-driven discovery of multi-system backends from intent, is what DDS targets. Infrastructure-as-code with LLMs (Pulumi-AI~\citep{pulumiai}, Terraform-AI~\citep{terraformmcp}, IaC-Eval~\citep{kon2024iaceval}, MACOG~\citep{khan2025macog}) renders a chosen architecture within one IaC dialect; we sit above that, with any such IaC backend a physical-layer implementation under our L3--L4 contracts (extended treatment in Appendix~\ref{app:extended-related}).

\paragraph{Composition demands typed contracts, and composition knowledge demands persistent memory.}
User intent carries constraints an agent cannot reliably infer unaided, along six typed dimensions (data model, access pattern, scale, latency, consistency, and cost; elaborated in \S\ref{subsec:intent}) that mirror the textbook view of a data-intensive application~\citep{kleppmann2017ddia}. ``One size fits all'' is settled~\citep{stonebraker2005onesize}, so every realistic intent forces composition. What agents need is not more context but typed contracts that bound the search: a validated intent, a type-checked operator DAG, skill contracts per system, and layer-attributed runtime signals. Empirical work on agent reliability supports this~\citep{cemri2025mast,pappu2026multiagentteams}: typed failures should route to the layer owning the violated decision rather than rely on free-form coordination. Composition knowledge is not just absent but unstable; connector configurations, recommended images, and version-specific quirks change with every release. A framework that hard-codes composition rules ages out within a release; a framework whose composition knowledge lives in per-system agent skills (\S\ref{subsec:skills}), edited from typed runtime signals, keeps pace, as the learning-loop (\S\ref{subsec:attribution-and-learning}) shows.

\section{The DDS Framework: Typed Contracts for Structured Agentic Discovery}
\label{sec:framework}

\begin{figure}[t]
\centering
\includegraphics[width=0.95\linewidth]{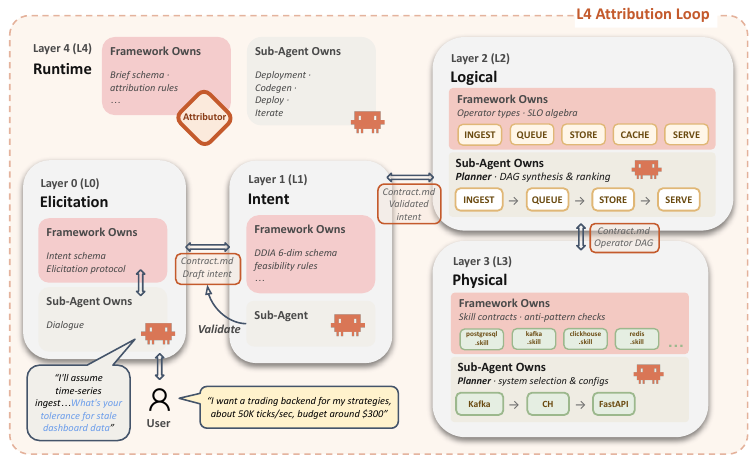}
\caption{The four DDS layers (L1--L4), each carrying a typed contract. L0 in the figure is an elicitation sub-step within L1: it produces the draft intent that the L1 contract validates, without owning its own contract. Framework-owned contracts sit above each layer; sub-agents do the work inside. L4 routes runtime signals to the layer owning the violated decision.}
\label{fig:layers}
\end{figure}

\textbf{Declarative Data Services (DDS)} is the structured-discovery framework sketched in \S\ref{sec:intro}. Four layers (L1--L4; Fig.~\ref{fig:layers}) each carry a typed contract. The L1 contract begins with an elicitation sub-step (depicted as L0 in Fig.~\ref{fig:layers}) in which a sub-agent produces the draft intent that the L1 contract then validates; the sub-step owns no separate contract because its output is the draft L1 commits to. Host-environment policy is a sub-policy at L4 rather than a separate layer. The framework owns the contract at each of L1--L4, sub-agents search inside each typed sub-space, and an attribution loop at the top routes every runtime signal back to the layer that owns the violated decision. The four sub-searches discover four different objects: a validated typed intent at L1, an SLO-feasible operator topology at L2, a product composition that fits the topology and clears anti-patterns at L3, and a deployable artifact at L4 (with skill patches accumulating across deployments as the loop runs). The framework makes these sub-searches compose into one deployment that meets original intents.

\textbf{The framework/agent split (the two ownership rules).} A pure rule-based framework cannot keep up with the breadth of product-specific knowledge a real stack requires, since every system has its own dialect, connector, and operational quirks. A pure prompt-to-code agent cannot enforce cross-layer contracts or trace a failure to the decision that caused it. DDS therefore commits to two ownership rules that hold at every layer L1--L4: \emph{(R1) the framework owns the contract}, meaning its types, schemas, composition rules, and validation, including attribution of runtime signals to the layer that owns the violated decision; \emph{(R2) the sub-agent owns the bounded search}, meaning the unconstrained, knowledge-intensive work that fits inside the contract (intent elicitation, DAG synthesis, product selection, codegen and deployment). The single global problem (``build me a working multi-system backend that meets this intent'') decomposes into four sub-searches over typed spaces, with the framework guaranteeing that the sub-searches compose. Each layer boundary carries a typed artifact (Fig.~\ref{fig:layers}) with three properties: it is \emph{validatable} (the framework rejects malformed inputs before codegen), \emph{citable} (downstream sub-agents quote specific fields in generated output), and \emph{editable} (a small change at one layer propagates downstream without rewriting code). Together these properties let knowledge pass forward through the layers as the search progresses, and let an L4 runtime signal land as a skill edit at L3 with no code change.

\phantomsection\label{subsec:intent}
\textbf{L1: intent contract.} Natural language is ambiguous; code and config commit to a product before the user has stated the requirement. An intent specification sits in between: a typed declaration over six dimensions (data model, access pattern, scale, latency, consistency, and cost). The framework checks well-formedness before any sub-agent writes code. An elicitation sub-agent (the L0 step within L1; Fig.~\ref{fig:layers}) produces the draft intent; the typed declaration is common ground between a non-expert user and downstream sub-agents.

\textbf{L2: operator DAG.} An agent can in principle jump from intent (``low-latency analytics over a stream'') straight to product selection (``use ClickHouse and Kafka'') in one prompt, but doing so collapses two distinct decisions: \emph{what topology} meets the workload, and \emph{which products} fill each role. A bad topology choice then invalidates every downstream product choice, and there is no clean way later to attribute a runtime failure to the topology versus the chosen system. The design decision at L2 is to separate these by committing to topology before products: a typed operator graph over an open set (\texttt{INGEST}, \texttt{STORE}, \texttt{TRANSFORM}, \texttt{SERVE}, \texttt{CACHE}, \texttt{QUEUE}, plus domain-specific extensions like chat's \texttt{ROUTE}, \texttt{NOTIFY}, \texttt{INDEX}; Appendix~\ref{app:chat}). The type system enforces three things the unstructured search cannot. (i) Every declared access pattern has a query path: reachability from each \texttt{INGEST} to every declared \texttt{SERVE} is a contract, not a hope. (ii) Per-edge guarantees compose into end-to-end SLOs, and a DAG whose aggregates miss the L1 budget is rejected before any codegen happens, so pattern alternatives surface here as ranked candidates rather than as silent agent choices buried in generated code. (iii) Pattern stays product-agnostic, which is what lets skills at L3 evolve independently and what lets later runtime signals attribute cleanly to L2 (the topology cannot meet the SLO) versus L3 (this product cannot meet the SLO under this configuration). The current SLO algebra is small and conservative (path latency sums, throughput minimums, consistency degrades to the weakest link); richer rules are open work (Appendix~\ref{app:open-problems}).

\phantomsection\label{subsec:skills}
\textbf{L3: skill contract.} The composition knowledge that distinguishes a working multi-system stack from a broken one---which connector, which configurations actually matter, and when \emph{not} to use a system---is scarce in pretraining: vendor docs do not document non-fit, and anti-patterns live in incident reports and team discussions. This knowledge also changes with every release cycle, so a pretraining-only agent ages out fast, and any fix expressed in a prompt evaporates with that prompt. The design decision at L3 is to make composition knowledge a first-class, persistent artifact of the framework rather than a transient instruction: an \emph{agent skill} is a structured, reusable, editable artifact, one per system, with four blocks (today materialized as YAML; Appendix~\ref{app:skill-yaml}): \texttt{capabilities}, \texttt{compositions}, \texttt{anti\_patterns}, and \texttt{operational}. The four-block split aligns to change rates: \texttt{capabilities} ages slowly with the system itself, while the other three change at the rate of release cycles, incidents, and host-environment churn, and are exactly the targets of runtime-driven patches. Two properties make L3 the \emph{learning unit} of DDS. (i) Persistent memory: a runtime signal can land as a skill edit, and the next deployment cites the patched line, so a fix made once carries forward to subsequent deployments of that system rather than evaporating with the prompt. (ii) Inline traceability: each non-obvious config decision in the deployed artifact is annotated downstream with the skill field that informed it, so the artifact carries the rule it was made to satisfy and a reviewer can audit each choice without rerunning the agent. The L3 planner sub-agent uses this catalog to filter and rank candidate products against the validated DAG; anti-pattern entries carry machine-checkable structured fields (\texttt{severity}, hard versus soft, plus matchers such as forbidden version ranges, type-incompatible columns, or known-bad operator pairings) that the framework enforces during planning, so a hard anti-pattern eliminates a candidate before any code is generated.

\begin{figure}[t]
\centering
\includegraphics[width=0.95\linewidth]{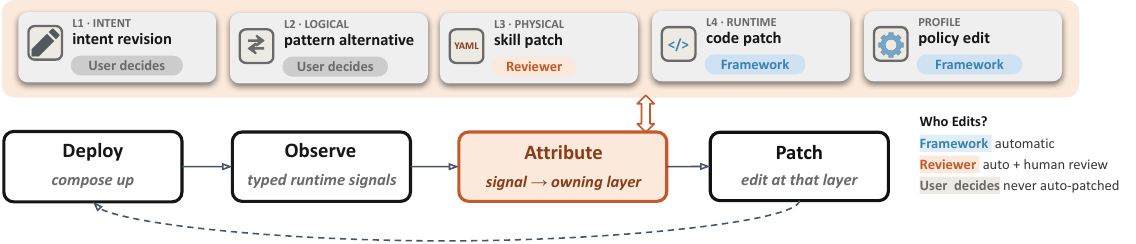}
\caption{The L4 attribution loop: deploy, observe, attribute, patch. Each runtime signal is routed to the layer that owns the violated decision; the edit unit and policy (framework auto, reviewer-in-the-loop, or user-decided) are set by that layer.}
\label{fig:attribution}
\end{figure}

\textbf{L4: attribution loop.} L4 deployment is the only layer that observes the running stack, so every runtime signal first lands here. The design decision is to \emph{type} each signal and route it back to the layer that owns the violated decision, rather than treat every failure as a generic agent error to retry. This is the difference between structured and unbounded discovery: without attribution, a failure reads ``the agent messed up,'' the next iteration searches the global space again, and there is no traceable connection between the symptom and the decision that caused it; with attribution, a failure becomes a bounded edit at one layer, and the next iteration searches only the affected sub-space. The routing (Table~\ref{tab:attribution}, Fig.~\ref{fig:attribution}) pairs each signal class with a correction policy matched to who can sensibly act: the framework auto-patches what it owns (transient codegen errors, host-environment policy entries); reviewer-in-the-loop edits land at L3 as skill patches that carry forward to every future deployment; L1 intent and L2 pattern edits are surfaced to the user rather than auto-applied because they cross a contract boundary the framework cannot revise. The classifier today is rule-based over deployment outputs (compose stderr, container exit codes and health-check states, container logs, smoke-verifier output); empirical evaluation is reported in \S\ref{subsec:attribution-and-learning}, and out-of-class behavior and a learned alternative to the rule-based classifier are future work (Appendix~\ref{app:open-problems}).

\begin{table}[t]
\centering
\caption{Runtime signals are typed and routed to the layer that owns the violated decision (cf.\ Fig.~\ref{fig:attribution}). Concrete signals from the trading case study and the skill patches that close them are in \S\ref{subsec:attribution-and-learning}. Some signals (consumer lag, p99 violation) are genuinely ambiguous between L2 and L3; the harness accepts either label and the open problem of a confidence model is discussed in Appendix~\ref{app:open-problems}.}
\label{tab:attribution}
\small
\begin{tabular}{@{}>{\raggedright\arraybackslash}p{0.20\linewidth}>{\raggedright\arraybackslash}p{0.30\linewidth}>{\raggedright\arraybackslash}p{0.09\linewidth}>{\raggedright\arraybackslash}p{0.30\linewidth}@{}}
\toprule
\textbf{Signal category} & \textbf{Examples} & \textbf{Layer} & \textbf{Edit (who applies)} \\
\midrule
Infeasible intent         & contradictory dimensions; budget vs.\ scale         & L1      & intent revision (user) \\
Pattern--SLO mismatch     & consumer lag; p99 violation                         & L2 / L3 (ambig.) & plan alternative or skill patch \\
Composition gap           & missing image; missing library                      & L3      & skill patch (reviewer) \\
Codegen slip              & transient compile or config error                   & L4      & code patch (framework, auto) \\
Host-environment mismatch & port conflict; missing host package                 & L4 (host) & policy entry (framework, auto) \\
\bottomrule
\end{tabular}
\end{table}

\section{Case Study: Building a Trading Backend with DDS}
\label{sec:case-study}

\textbf{Setup.}
The case study builds an end-to-end data backend for a solo-trader analytics workload from a single intent specification 
covering streaming market-data ingest, time-series aggregation, low-latency lookups, and a hobbyist budget; 
the full intent is in Appendix Fig.~\ref{fig:intent-yaml}. We evaluate each generated deployment at three tiered levels 
(T3: declared SLOs hold under load (deferred to \S\ref{sec:discussion})):
\begin{itemize}
\itemsep0pt
\item \textbf{T0:} all generated artifacts have valid syntax;
\item \textbf{T1:} the stack boots to steady state under \texttt{docker compose up} and all healthchecks pass;
\item \textbf{T2:} a smoke query returns expected rows end-to-end.
\end{itemize}
We report three things: an end-to-end DDS walkthrough (\S\ref{subsec:walkthrough}); a headline comparison of unbounded-discovery baselines against DDS plus a skill-content ablation that isolates which parts of the skill artifact carry the win (\S\ref{subsec:headline}); and the L4 attribution loop on real failures and controlled fault injection (\S\ref{subsec:attribution-and-learning}). All runs use the same Claude model (\texttt{claude-opus-4-6}) and tool access; only the framework surface and iteration regime vary across conditions. A second case study on a chat platform that stresses operator-algebra extensibility is in Appendix~\ref{app:chat}.

\subsection{End-to-end walk: intent to live data}
\label{subsec:walkthrough}

\textbf{Intent through deployment.}
The elicitation sub-agent produces a typed intent that populates all six L1 dimensions; 
framework validation at L1 emits one soft warning (cost preference under-specified, defaulted 
to \emph{simplicity}) and no hard errors. The planner sub-agent at L2 synthesizes the operator 
DAG: \texttt{INGEST}~$\to$~\texttt{QUEUE}~$\to$~\texttt{TRANSFORM}~$\to$~\{\texttt{STORE}(analytics), 
\texttt{STORE}(operational), \texttt{CACHE}\}, and the framework type-checks each edge and the 
SLO composability of each path. At L3, the planner sub-agent selects 
Kafka (QUEUE), ClickHouse (STORE analytics, with Kafka Engine and an OHLCV materialized view), 
PostgreSQL (STORE operational, positions table), and Redis (CACHE, hot state). 
The framework validates each candidate against the skill's anti-patterns and each pair 
against the adjacent skill's composition rules; the plan fits the declared \$300/mo envelope. 
At L4, the deployment sub-agent receives a structured brief (what to generate, which skill fields to cite,
what checks must pass), emits the artifacts in Table~\ref{tab:artifacts}, and runs \texttt{docker compose up -d}.

\begin{figure}[t]
\centering
\includegraphics[width=0.85\linewidth]{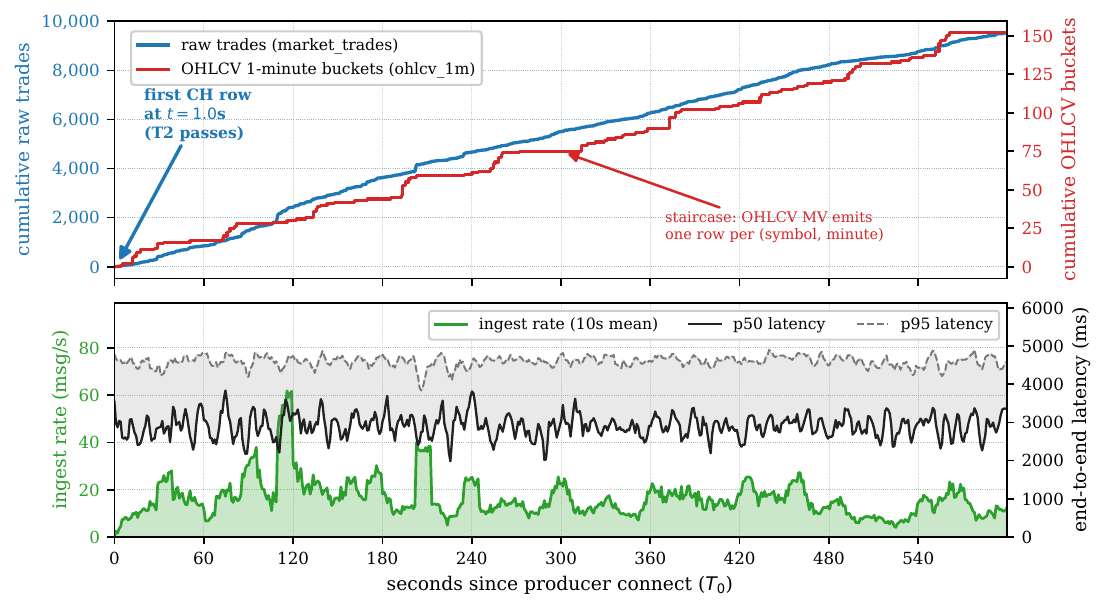}
\caption{Live-data proof of life on a DDS-generated stack. A Coinbase public WebSocket feed (20 USD pairs) is plugged into the running stack at $T_0$. 
\textbf{Top:} cumulative raw trades (blue) and OHLCV 1-minute buckets (red); the red staircase is the \texttt{ohlcv\_1m} 
materialized view emitting one row per (symbol, minute). \textbf{Bottom:} 10-second rolling ingest rate (green) tracks the exchange's bursty traffic; 
end-to-end latency p50 (black) and p95 (dashed) stay bounded across the full window.}
\label{fig:trading-timeline}
\end{figure}

\textbf{Boot, smoke, and live data.}
A single representative DDS one-shot run (one of the 10 reported in Appendix~\ref{app:e8}) passes T0, T1, and T2 (the smoke query returns end-to-end rows); the generated deployment comprises roughly 1{,}100 lines across six services. To show that a T1-passing stack is not a Potemkin deployment, Fig.~\ref{fig:trading-timeline} reports one 10-min window in which a live public exchange feed (Coinbase spot, 20 USD pairs) is plugged into a DDS-generated stack at the Kafka ingress. (The L4 sub-agent generated Binance-targeted producers in Table~\ref{tab:artifacts}; for the proof of life we substitute Coinbase because its public feed needs no authentication.) Trades land in the ClickHouse raw table within 1~s of producer connect; the OHLCV materialized view emits one row per (symbol, minute); ingest tracks the exchange's bursty traffic (mean 15.9~msg/s); end-to-end latency from exchange timestamp to ClickHouse-observable row holds at roughly 3~s p50 / 4.7~s p95 across the window. Nothing is exercised beyond what the DDS-planned topology already provides. We report this run as a proof of life, not a scale test: the 10-min window runs at a single-exchange public-feed rate (mean 15.9~msg/s), below the declared sustained target of 100 events/sec in the intent (Fig.~\ref{fig:intent-yaml}, representing multi-exchange aggregation); validating declared SLOs under sustained load (T3), together with correctness and resilience probes, is deferred to the L5 evaluator (\S\ref{sec:discussion}, Appendix~\ref{app:l5}).

\begin{table}[t]
\centering
\caption{Artifacts generated by a representative DDS run (T0, T1, T2 all pass).}
\label{tab:artifacts}
\footnotesize
\setlength{\tabcolsep}{4pt}
\begin{tabular}{lllr}
\toprule
\textbf{Artifact} & \textbf{File} & \textbf{Purpose} & \textbf{LoC} \\
\midrule
Docker Compose           & \texttt{docker-compose.yml}                      & Six-service stack with healthchecks             & 129 \\
ClickHouse DDL           & \texttt{clickhouse/init.sql}                     & Kafka Engine + OHLCV materialized view          & 244 \\
PostgreSQL DDL           & \texttt{postgresql/init.sql}                     & Positions table, indexes                        & 127 \\
Producer (Python)        & \texttt{binance\_market\_stream.py}              & Exchange-to-Kafka market-data ingest            & 131 \\
Producer (Python)        & \texttt{binance\_user\_data\_stream.py}          & Exchange-to-Kafka user-data ingest              & 165 \\
Producer (Python)        & \texttt{order\_book\_snapshots\_top20.py}        & Order-book snapshot ingest                      & 144 \\
Smoke verifier           & \texttt{verify.py}                               & End-to-end T2 check                             & 148 \\
\midrule
\textbf{Total}           &                                                  &                                                 & \textbf{1{,}088} \\
\bottomrule
\end{tabular}
\end{table}

\begin{center}
\begin{takeawaybox}{Structured Discovery Yields a Working Backend in One Cycle}
Starting from a user-declared intent, DDS searches the four typed sub-spaces in a single deployment cycle, producing ${\sim}1{,}100$ lines across six services that boot to steady state, pass the end-to-end smoke query, and sustain a live exchange feed for 10~min.
\end{takeawaybox}
\end{center}

\subsection{Where the win comes from}
\label{subsec:headline}

\textbf{DDS converges consistently; iterated raw agents do not.}
We compare four conditions on the same trading intent, using the same Claude model and the same tool access (Table~\ref{tab:headline}). Conditions A, B, and C are versions of \emph{unbounded agentic discovery} with progressively more knowledge in the prompt; DDS is \emph{structured agentic discovery} with typed contracts at all four layers. All four conditions use the same 5-iteration outer feedback loop modeled on real-world Claude-Code usage: after each codegen attempt, the harness runs T0/T1/T2 acceptance and feeds the failure log back to the agent for editing in the next iteration. This eliminates the trivially-fixable single-shot failure modes (host-port conflicts, missing init files) and tests whether iteration alone is sufficient for a determined user without a framework. Conditions vary only in iter-1 prompt content: A is a natural-language requirements prompt with no system guidance (free-form search over systems and topology); B adds the explicit list of systems to use (Kafka, ClickHouse, PostgreSQL, Redis), narrowing the system catalog; C additionally pastes the four DDS skill YAMLs as prose into the prompt, loading composition knowledge into context without typed channels for it to flow forward (knowledge-loaded but still unbounded). \textbf{Condition A} ($n{=}10$) reached T1 2/10 at median \$5.76, with median 268 total turns across the 5 iterations; six of eight T1 failures hit \texttt{iterations\_exhausted} after 5 rounds (per-run trajectories in Appendix~\ref{app:failure-analysis}). \textbf{Condition B} ($n{=}10$) reached T1 3/10 at median \$6.23 with median 209 turns: telling the agent which systems to use marginally improves boot rate but does not close the gap. \textbf{Condition C} ($n{=}10$) reached T1 6/10 and T2 4/10 at median \$5.06 with median 229 turns: pasting the four skill YAMLs as prose closes most of the rate gap relative to A and B, confirming that skill content carries real value, but at $\sim$5$\times$ the median turn cost of DDS+iter. \textbf{DDS + 5-iter} ($n{=}10$, same outer feedback loop as A/B/C) reached T0, T1, and T2 at \textbf{10/10, 10/10, 10/10} at median \$1.94 with median 44 turns; every run terminated with \texttt{stop\_reason=passed}, all ten by iter 4 with median 2 iterations to T1. Same model, same tools, same iteration scope; the delta is the contract surface that focuses each iteration's edits, not iteration itself. For reference, DDS without the outer loop already reached T1 8/10 at \$1.49 in median 17.5 turns (Appendix~\ref{app:e8}): contracts alone do most of the work; the outer loop closes the remaining T2 gap.

\textbf{Why unbounded discovery fails to converge.}
Two structural patterns recur across the unbounded-discovery baselines (per-run trajectories in Appendix~\ref{app:failure-analysis}): without an L2 pin on topology, a downstream failure can trigger an architectural rewrite that regresses an earlier-passing tier (the search re-enters the global space rather than the local one); and when the failure log surfaces a symptom without identifying a layer, the agent guesses at the wrong sub-search and iteration loops without progress. Structured discovery prevents both, because L1/L2 fix the topology before any codegen and L4 attribution routes each typed signal to the layer that owns the violated decision.

\textbf{The outer loop closes the smoke-timing gap.}
Without an outer T1/T2 retry, DDS one-shot reaches T2 only 3/10, and the five T1-but-not-T2 runs share one root cause: materialized-view population timing combined with agent non-determinism in MV column naming. The 5-iter outer loop is a separate $n{=}10$ batch (Appendix~\ref{app:dds-iter-perrun}); in that batch every run reaches T2 at iter 4 or earlier, and the two failure classes that drag T2 down in one-shot ($\texttt{smoke\_query\_error}$, $\texttt{smoke\_empty}$) close after at most one extra iteration. The first-iteration T1 rate in the iterated batch (2/10) is lower than one-shot's 8/10 under nominally the same model and tools, a sampling gap at $n{=}10$ that a larger sample would resolve. This is the L4 attribution loop compressed to a single run: a typed runtime signal at iteration $i$ becomes a focused L4 edit at iteration $i{+}1$, no skill patch required.

\begin{table}[t]
\centering
\caption{Headline comparison on the trading intent. Same Claude model and tool access throughout. A, B, and C use a 5-iteration outer feedback loop (T0/T1/T2 acceptance, failure-log fed to the next iter; max 5 iters/run). The two DDS rows isolate that outer loop: \emph{one-shot L4} uses the internal T0 acceptance loop only; \emph{+ 5-iter loop} adds the same outer loop. Per-run detail in Appendices~\ref{app:failure-analysis} (A), \ref{app:e8} (DDS one-shot), and \ref{app:dds-iter-perrun} (DDS + 5-iter).}
\label{tab:headline}
\small
\begin{tabular}{lccccc}
\toprule
\textbf{Condition} & \textbf{n} & \textbf{T0} & \textbf{T1} & \textbf{T2} & \textbf{Median \$} \\
\midrule
A: NL only + 5-iter loop                  & 10 & 10/10 & 2/10  & 2/10  & \$5.76 \\
B: NL + system names + 5-iter loop        & 10 & 10/10 & 3/10  & 2/10  & \$6.23 \\
C: + skill YAMLs (prose) + 5-iter loop    & 10 & 10/10 & 6/10  & 4/10  & \$5.06 \\
\midrule
DDS: full framework, one-shot L4          & 10 & 9/10  & 8/10  & 3/10           & \$1.49 \\
DDS: full framework + 5-iter loop         & 10 & 10/10 & \textbf{10/10} & \textbf{10/10} & \$1.94 \\
\bottomrule
\end{tabular}
\end{table}

\textbf{Inside the skill contract: operational and anti-pattern content does the heavy lifting.}
The skill ablation holds the framework and intent fixed and varies the content of the agent skills across three settings, $n{=}3$ runs per variant (Table~\ref{tab:ablation}). The full variant (all four blocks) reaches T1 2/3 at median \$1.36. Ops-stripped (drop \texttt{operational} and \texttt{anti\_patterns}, keep \texttt{capabilities} and \texttt{compositions}) drops to 0/3 at \$1.11. Minimal (keep only \texttt{capabilities}) is also 0/3 at \$0.65. Code-quality probes localize the damage: dropping the operational block removes the Kafka recommended-image field (full: 3/3 use the recommended image; ops-stripped: 0/3) and the PostgreSQL host-port-conflict remap (full: 3/3 remap; ops-stripped: 0/3). These are exactly the fields whose patches close the first-deployment failures in \S\ref{subsec:attribution-and-learning}. Composition content helps a little; capabilities content is largely redundant with pretraining. Cost rises with skill content because a more complete skill drives a more complete artifact: the full variant emits the entire stack and pays for it, while stripped variants give up partway and produce nothing that boots. Sample size here is small ($n{=}3$); a larger replication is left to future work.

\begin{table}[t]
\centering
\caption{Skill-content ablation: framework, intent, and tools held constant; only the contents of the agent skills vary. Single-shot DDS (no outer iteration loop), $n{=}3$ runs per variant. Operational and anti-pattern content does the heavy lifting; capabilities content is largely redundant with pretraining.}
\label{tab:ablation}
\small
\begin{tabular}{lccccc}
\toprule
\textbf{Skill variant} & \textbf{n} & \textbf{T0} & \textbf{T1} & \textbf{T2} & \textbf{Median \$} \\
\midrule
full (all four blocks)      & 3  & 3/3  & 2/3  & 2/3  & \$1.36 \\
ops-stripped (drop \texttt{operational} + \texttt{anti\_patterns}) & 3  & 3/3  & 0/3  & 0/3  & \$1.11 \\
minimal (\texttt{capabilities} only)                & 3  & 3/3  & 0/3  & 0/3  & \$0.65 \\
\bottomrule
\end{tabular}
\end{table}

\begin{center}
\begin{takeawaybox}{Structure Carries the Win, Not Iteration}
The same skill content wins inside DDS and loses inside an unbounded prompt: the typed channel, not the knowledge, is what turns iteration into a layer-bounded edit.
\end{takeawaybox}
\end{center}

\subsection{Closing the L4 loop}
\label{subsec:attribution-and-learning}

\textbf{Attribution closes the loop on the in-distribution fault classes that arise in practice.}
The L4 pipeline (compose-stderr and container-log signal collection, rule-based classification, layer routing) is exercised on five fault classes drawn from this case study: image-pull failure, host-port conflict, library-missing, and DDL-type constraint match F1--F4 in Table~\ref{tab:failures}; consumer-lag is the canonical L2-versus-L3 ambiguous case from Table~\ref{tab:attribution} (the harness accepts either label). Across four controlled injections per class ($n{=}20$), each pairing a known-cause fault with a known-good artifact, the predicted layer matched ground truth on every instance and the routing mirrored the uncontrolled F1--F4 outcome: every signal landed where its skill patch was applied. This is in-distribution behavior on the classes the rules were authored against (a sanity check, not a generalization claim); out-of-class behavior and learned attribution are the central open problems (Appendix~\ref{app:open-problems}).

\textbf{Real failures become skill patches, not prompts to retry.}
The learning-loop experiment ran DDS twice on the trading intent. The first deployment surfaced four distinct runtime failures (Table~\ref{tab:failures}), each attributed and patched accordingly. Three failures labeled L3 skill-gaps: a missing recommended image for Kafka; missing Python extras for the producer; and a missing anti-pattern for \texttt{TTL} on \texttt{DateTime64} in ClickHouse. One labeled profile-level (host-port). Remediation consisted of skill-only patches to three skill files plus one profile entry; no code, no prompt, and no architecture changes. The second deployment fixed 4/4 with 0 regressions (Table~\ref{tab:failures}), and the patched fields are cited inline in the generated artifacts. One new failure of the same class as F2 surfaced during the second deployment in a different host environment and was patched the same way, showing that the loop absorbs fresh signals without code change.

\begin{table}[t]
\centering
\caption{First-deployment failures on the trading backend and their layer attribution. All four observed at L4; three attribute to L3 skill gaps. The second deployment fixed 4/4 with skill patches.}
\label{tab:failures}
\footnotesize
\setlength{\tabcolsep}{4pt}
\begin{tabular}{lllll}
\toprule
\textbf{ID} & \textbf{Signal} & \textbf{Direct layer} & \textbf{Root cause} & \textbf{Artifact} \\
\midrule
F1 & image-pull-failure   & L4 & L3: missing \texttt{recommended\_images}            & skill patch \\
F2 & host-port-conflict   & L4 & profile and L3: no port probe                       & profile + skill patch \\
F3 & ddl-type-constraint  & L4 & L3: missing \texttt{TTL on DateTime64} anti-pattern & skill patch \\
F4 & library-missing      & L4 & L3: missing \texttt{required\_python\_extras}       & skill patch \\
\bottomrule
\end{tabular}
\end{table}

\begin{center}
\begin{takeawaybox}{Runtime Failures Land as Patches}
Four real first-deployment failures (F1--F4) attribute to L3 or profile; 
skill-only edits fix 4/4 on the second deployment with no code or prompt change, 
and the patched fields appear as inline citations in the next deployment's artifacts. 
\end{takeawaybox}
\end{center}

\FloatBarrier

\section{Discussion and Open Problems}
\label{sec:discussion}

\textbf{Roadmap: an L5 evaluator peer layer.}
The most important next step is an \textbf{evaluator} peer layer (L5) that would ask whether a deployed stack actually meets the declared intent. The L4 loop today carries only crash-time signals; L5 would surface SLO-time and correctness-time signals through the same layer-attributed routing, and would make T3 (declared SLOs holding under load) the first acceptance tier the framework makes routine. The same framework/agent split would apply: the framework would own probe dimensions, access, and attribution; the sub-agent would synthesize intent-specific semantic probes (OHLCV bar continuity in trading, message-ordering preservation in chat) that no fixed harness ships. L5 is the natural answer to validation without a ground-truth oracle: it would convert ``is the deployed stack correct?'' from an unanswerable question into a structured search over typed probes, with attribution back to the layer that owns the violated decision. Detail, the broader case for verifier-driven systems work, and other roadmap axes (operator-algebra inference at L2, a cross-product cost objective, semi-automatic skill extraction, security/governance) are in Appendix~\ref{app:l5}.
\textbf{Open challenges this prototype does not yet solve.}
The quantitative headline is one domain (trading), one model (Claude Opus 4.6), and one host environment; a second domain (ecommerce, $n{=}5$) replicates T0--T3 under the same framework, and DDS-Bench is built to broaden this coverage (Appendix~\ref{app:bench-scope}), but the chat case study remains descriptive. The most direct extension of Table~\ref{tab:headline} is re-running Condition C on a non-Anthropic model to separate the structure-vs-iteration claim from model-specific behavior; this is in flight. The $n{=}20$ attribution check in \S\ref{subsec:attribution-and-learning} is in-distribution over the rule-authored classes; out-of-class behavior and learned attribution are open (Appendix~\ref{app:open-problems}). The prototype runs on local Docker Compose, and the proof-of-life sustains a single-exchange feed (mean 15.9 msg/s) below the declared 100 events/sec target in Fig.~\ref{fig:intent-yaml}; production-grade concerns and the load gap each map to a specific DDS layer when taken on (Appendix~\ref{app:l5}). Other open problems span the four layers: attribution confidence at L4, composition-rule inference and SLO-algebra extension at L2, intent drift and elicitation protocol at L1, and skill versioning at L3 (Appendix~\ref{app:open-problems}).

\section{Related Work}
\label{sec:related}

Adjacent directions on the data-systems side are in \S\ref{sec:motivation}; extended treatment of adjacent work, including the agentic-discovery sibling line, is in Appendix~\ref{app:extended-related}.

\bibliographystyle{unsrtnat}
\bibliography{references}

\appendix

\section{Example Agent Skill: ClickHouse}
\label{app:skill-yaml}

Figure~\ref{fig:skill-yaml} shows a trimmed excerpt of \texttt{clickhouse.yaml} skill with one representative entry per block. The dated comments are real attribution-log entries: each was added after a specific failure during the learning-loop experiment (\S\ref{subsec:attribution-and-learning}), which is the traceability property cited in \S\ref{subsec:skills}.

\begin{figure}[h!]
\centering
\begin{minipage}{0.92\linewidth}
\caption{Trimmed \texttt{skills/clickhouse.yaml} showing one entry per block (capabilities, operational, anti\_patterns, compositions). Dated comments are real attribution-log entries from the learning-loop experiment.}
\small
\begin{verbatim}
skill:
  system: clickhouse
  version: "24.3"
  operator_types: [STORE, TRANSFORM]

  capabilities:
    data_models: [columnar, time_series, event]
    access_patterns: [olap, streaming]
    max_throughput: "500K inserts/sec per node"
    consistency: [eventual]

  operational:
    recommended_images:
      - "clickhouse/clickhouse-server:24.3"
    # CH exposes 9000 (native TCP) and 8123 (HTTP); 9000 is a very
    # common dev-host collision target.
    known_host_port_conflicts:
      - port: 9000
        remap_to: 19000
        reason: "CH native TCP port 9000 commonly occupied on dev hosts"

  anti_patterns:
    - scenario: "TTL expression on a DateTime64 column (CH <=24.x)"
      reason: "TTL requires DateTime or Date; DateTime64 not accepted directly"
      alternative: "Wrap: TTL toDateTime(event_time) + INTERVAL 6 MONTH"
      severity: hard_limit

    - scenario: "OLTP point updates (UPDATE/DELETE by primary key)"
      reason: "MergeTree is append-only; mutations are async background ops"
      alternative: "PostgreSQL for OLTP workloads"
      severity: hard_limit

  compositions:
    - with: kafka
      connector: kafka_engine_materialized_view
      direction: inbound
      semantics: at_least_once
      known_issues:
        - "Kafka Engine creates a virtual table; use MaterializedView to persist"
\end{verbatim}
\end{minipage}
\label{fig:skill-yaml}
\end{figure}

\section{Per-run detail for DDS one-shot (no outer loop)}
\label{app:e8}

Table~\ref{tab:e8-perrun} reports every run of the $n{=}10$ DDS one-shot configuration \emph{without} the outer 5-iter feedback loop used in Table~\ref{tab:headline}. The headline DDS row uses the outer loop (T0/T1/T2 at 10/10, 10/10, 10/10 at \$1.94, 44 turns); this appendix shows what DDS reaches with the contract surface alone (T1 8/10, T2 3/10, median \$1.49, median 17.5 turns). All runs use the same intent, same model, same tool access, and the canonical skills directory.

\begin{table}[h]
\centering
\caption{Per-run detail for the $n{=}10$ DDS one-shot configuration (no outer loop).}
\small
\begin{tabular}{rcccrrrc}
\toprule
\textbf{Run} & \textbf{T0} & \textbf{T1} & \textbf{T2} & \textbf{Turns} & \textbf{Tokens} & \textbf{Cost (\$)} & \textbf{Signals} \\
\midrule
1  & Y & Y & N & 15 & 43{,}330 & 1.53 & 2 \\
2  & Y & N & N & 33 & 41{,}890 & 1.64 & 1 \\
3  & Y & Y & N & 18 & 32{,}957 & 1.19 & 1 \\
4  & Y & Y & Y & 17 & 40{,}852 & 1.44 & 0 \\
5  & N & Y & N & 15 & 33{,}830 & 0.91 & 0 \\
6  & Y & Y & Y & 19 & 43{,}417 & 1.66 & 0 \\
7  & Y & Y & N & 16 & 48{,}297 & 1.64 & 0 \\
8  & Y & Y & N & 14 & 38{,}126 & 1.22 & 1 \\
9  & Y & Y & Y & 18 & 35{,}940 & 1.24 & 0 \\
10 & Y & N & N & 20 & 51{,}436 & 2.12 & 1 \\
\midrule
\textbf{Total} & \textbf{9/10} & \textbf{8/10} & \textbf{3/10} & median 17.5 & median 41{,}371 & median 1.49 & \\
\bottomrule
\end{tabular}
\label{tab:e8-perrun}
\end{table}

The two T1 failures are diagnosable: run 2 hit a boot-failure signal at L4 that auto-repair did not clear within the turn budget, and run 10 hit a host-port conflict (the same fault class as the first-deployment failure F2 in \S\ref{subsec:attribution-and-learning}) on a host whose port policy was not yet captured in the profile. The five T1-but-not-T2 runs (1, 3, 5, 7, 8) break down as 3 \texttt{smoke\_query\_error} and 2 unsignalled (runs 5 and 7, with 0 typed signals in the table), all consistent with materialized-view-population timing combined with agent non-determinism in MV column naming: the stacks are running, and the smoke probe window does not account for producer-priming variance. Run 5 reports T0~$=$~N with T1~$=$~Y because the T0 syntax check flagged a non-blocking script (the \texttt{verify.py} smoke verifier in Table~\ref{tab:artifacts}) rather than a service artifact; the six container artifacts were syntactically valid, so the stack still booted to steady state and passed T1 (the smoke probe itself failed at T2, consistent with the same root cause as the other T1-but-not-T2 runs). The DDS-with-outer-loop variant in Appendix~\ref{app:dds-iter-perrun} is a separate $n{=}10$ batch in which the same root cause closes after at most one extra iteration, taking T2 from 3/10 (one-shot) to 10/10 (with outer loop).

\section{Per-run detail for DDS with 5-iter outer loop}
\label{app:dds-iter-perrun}

Table~\ref{tab:dds-iter-perrun} reports every run of the $n{=}10$ DDS configuration that uses the same 5-iteration outer feedback loop as Conditions A/B/C in Table~\ref{tab:headline}. The L1/L2/L3 framework pipeline is computed once per run (deterministic for the fixed intent); only the L4 deployment sub-agent is re-invoked on T0/T1/T2 failures with the original brief plus the failure log from the previous iteration. All runs use the same intent, model, and tool access as Appendix~\ref{app:e8}; the only difference is the outer loop.

\begin{table}[h]
\centering
\caption{Per-run detail for the $n{=}10$ DDS + 5-iter outer-loop configuration. ``First$\to$T1'' is the iteration in which T1 first passed (smaller is better; iteration 1 = no outer feedback needed).}
\small
\begin{tabular}{rcccrrrrr}
\toprule
\textbf{Run} & \textbf{T0} & \textbf{T1} & \textbf{T2} & \textbf{Iters} & \textbf{First$\to$T1} & \textbf{Turns} & \textbf{Wall (s)} & \textbf{Cost (\$)} \\
\midrule
1  & Y & Y & Y & 3 & 3 & 83  & 1{,}121 & 2.25 \\
2  & Y & Y & Y & 1 & 1 & 32  & 684     & 1.51 \\
3  & Y & Y & Y & 4 & 4 & 124 & 1{,}455 & 3.39 \\
4  & Y & Y & Y & 2 & 2 & 33  & 796     & 1.58 \\
5  & Y & Y & Y & 2 & 2 & 33  & 1{,}005 & 2.08 \\
6  & Y & Y & Y & 2 & 2 & 49  & 823     & 1.72 \\
7  & Y & Y & Y & 4 & 4 & 141 & 1{,}628 & 4.10 \\
8  & Y & Y & Y & 4 & 3 & 99  & 853     & 1.81 \\
9  & Y & Y & Y & 2 & 2 & 40  & 1{,}070 & 2.23 \\
10 & Y & Y & Y & 1 & 1 & 16  & 574     & 1.11 \\
\midrule
\textbf{Total} & \textbf{10/10} & \textbf{10/10} & \textbf{10/10} & median 2 & median 2 & median 44.5 & median 929 & median 1.945 \\
\bottomrule
\end{tabular}
\label{tab:dds-iter-perrun}
\end{table}

Two patterns are worth noting. (i) Convergence is fast: 6 of 10 runs reached T1 within 2 iterations (2 of 10 in iter 1 with no outer feedback needed, plus 4 of 10 in iter 2); median is 2 iterations to T1, and all 10 close by iter 4 (2 runs each at iter 3 and iter 4). (ii) The outer loop's marginal cost over one-shot DDS is small: median turns rose from 17 (one-shot) to 44 (with outer loop), median cost from \$1.49 to \$1.94, while T1 went 8/10 $\to$ 10/10 and T2 went 3/10 $\to$ 10/10. Even the most expensive DDS+iter run (run 7 at 141 turns, \$4.10) is below the median cost of every iterated baseline (A: 268 turns/\$5.76; B: 209/\$6.23; C: 229/\$5.06).

\section{Second case study: a chat platform}
\label{app:chat}

To stress the operator algebra beyond the trading workload, we ran DDS on a Signal-like consumer chat platform. The original operator set (\texttt{INGEST}, \texttt{STORE}, \texttt{TRANSFORM}, \texttt{SERVE}, \texttt{CACHE}, \texttt{QUEUE}) was insufficient. Three new operators (\texttt{ROUTE}, \texttt{NOTIFY}, \texttt{INDEX}) were added to the framework's open operator-type set as a one-time framework edit between runs; the L2 planner sub-agent then used them automatically against the chat intent to handle message delivery, push notifications, and full-text search. The honest reading is therefore that the L2 type system is \emph{open} (it admits new operators without restructuring the contract), not that the planner sub-agent \emph{discovered} the new operators at planning time; a principled planner-side extension-proposal protocol, with composition-rule inference for the proposed operators, is named as open work in Appendix~\ref{app:open-problems}. The physical mapping chosen by the planner given the extended operator set: Kafka~$\to$~NATS~$\to$~ScyllaDB + PostgreSQL + Elasticsearch + S3. Two surprises shaped our view of what L2 should represent. First, security (end-to-end encryption) acts as an \emph{architectural force} that constrains which component can hold which key, rather than a checkbox attached to any single component; it is a first-class intent dimension that cross-cuts L1 and L3. Second, ephemeral versus persistent data (presence indicators, typing indicators) is an \emph{access-pattern} property rather than a store property, and belongs at L2. This run is descriptive rather than quantitative: its role is to stress algebra extensibility, which is the open-operator property at L2, and to surface dimensions the trading workload does not exercise.

\section{Per-run failure analysis for iterated baselines (A, B, C)}
\label{app:failure-analysis}

This appendix complements Table~\ref{tab:headline} with per-run trajectories from the iterated baselines. All three conditions use the same 5-iteration feedback loop modeled on real-world Claude-Code usage: after each codegen attempt, the harness runs T0/T1/T2 acceptance and feeds the failure log back to the agent for editing in the next iteration. Up to 5 iterations per run; no per-run cost or wall-clock cap. Conditions vary only in iter-1 prompt content (A: NL only; B: NL + system names; C: + skill YAMLs as prose).

\paragraph{Per-condition summary.}
\begin{center}
\small
\begin{tabular}{lcccccc}
\toprule
\textbf{Condition} & \textbf{n} & \textbf{T0} & \textbf{T1} & \textbf{T2} & \textbf{med.\ turns} & \textbf{med.\ \$} \\
\midrule
A: NL only + 5-iter loop                & 10 & 10/10 & 2/10  & 2/10  & 268 & \$5.76 \\
B: NL + system names + 5-iter loop      & 10 & 10/10 & 3/10  & 2/10  & 209 & \$6.23 \\
C: + skill YAMLs (prose) + 5-iter loop  & 10 & 10/10 & 6/10  & 4/10  & 229 & \$5.06 \\
DDS: full framework + 5-iter loop       & 10 & 10/10 & \textbf{10/10} & \textbf{10/10} & 44  & \$1.94 \\
\bottomrule
\end{tabular}
\end{center}

Across the 10 A runs, T0 passed in every case (eventually) but T1 closed only twice; six of the eight T1 failures hit \texttt{iterations\_exhausted} after 5 rounds with no convergence. Median total turns across A is 268 ($\sim$15$\times$ DDS's median 17.5 turns; \S\ref{subsec:headline}, Appendix~\ref{app:e8}). We annotate three failure trajectories that capture distinct ways iteration fails to close the gap, plus one success trajectory showing what convergence looks like under iteration alone.

\subsection*{Failure trajectory 1: port conflict resolved, then a different system gets stuck (run 1)}

\noindent
\textbf{Outcome:} 5 iterations, 249 turns, $\sim$28 min wall; T0 \checkmark, T1 \ding{55}, T2 \ding{55}.

\begin{itemize}[leftmargin=*]
\itemsep0pt
\item \textbf{Iter 1:} 48 turns, 0 in-scope edits (T0 \ding{55}). Agent explored without producing complete artifacts.
\item \textbf{Iter 2:} 60 turns, 14 in-scope edits (T0 \checkmark, T1 \ding{55}). Stack now boots far enough to hit \texttt{address already in use} on PostgreSQL host port 5432.
\item \textbf{Iter 3:} 45 turns, 2 in-scope edits. Agent remaps the PostgreSQL port. T1 still fails: \texttt{container redpanda is unhealthy}.
\item \textbf{Iter 4--5:} 38 + 58 turns, 1 + 1 in-scope edits each. The Redpanda failure persists; the agent guesses at config without resolving it.
\end{itemize}

\noindent
\textbf{Root cause.} The Redpanda container becomes unhealthy because the agent's chosen broker configuration is incomplete or wrong, but the failure feedback the harness pipes back is only \texttt{dependency failed to start: container redpanda is unhealthy}. Without seeing the broker's own log, the agent cannot diagnose which flag to fix; it edits adjacent configuration files (1--2 edits per iteration) hoping to influence the symptom. Five iterations and 249 turns yield no convergence.

\noindent
\textbf{How DDS prevents this.} The Kafka skill (Appendix~\ref{app:skill-yaml}) lists \texttt{recommended\_images: [apache/kafka:3.7.0, ...]} and skips Redpanda. The L3 planner therefore selects apache/kafka, the L4 brief requires citing the skill's \texttt{compositions[with: clickhouse]} connector pattern in the generated artifact, and \texttt{operational.known\_host\_port\_conflicts} pre-empts the 5432 conflict. DDS reaches T1 in a single deployment without iteration.

\subsection*{Failure trajectory 2: iteration regresses a working stack (run 5)}

\noindent
\textbf{Outcome:} 5 iterations, 273 turns, $\sim$47 min wall; T0 \checkmark, T1 \ding{55}, T2 \ding{55}.

\begin{itemize}[leftmargin=*]
\itemsep0pt
\item \textbf{Iter 1:} 36 turns, 20 in-scope edits. \emph{T0 \checkmark, T1 \checkmark}, T2 \ding{55}. The agent's first-shot TimescaleDB-based stack actually \emph{boots cleanly}; only the smoke query fails, plausibly a one-line fix.
\item \textbf{Iter 2:} 78 turns, 5 in-scope edits. The agent restructures the topology (adding Kafka and switching from TimescaleDB to a more elaborate stack) to ``fix'' T2. The new compose hits a container-name conflict because the old container from iter 1 was still present. T1 \ding{55}.
\item \textbf{Iter 3--5:} 27 + 78 + 54 turns. Container-name conflict resolved; now \texttt{kafka is unhealthy} across all three iterations. The agent never recovers the working topology it had in iter 1.
\end{itemize}

\noindent
\textbf{Root cause.} Iteration here is actively harmful: the agent's response to a T2 failure was an \emph{architectural rewrite}, not a focused fix. With no contract pinning the L2 topology, the agent re-derives the system selection at every iteration boundary, which can pivot the entire stack on a single feedback signal.

\noindent
\textbf{How DDS prevents this.} The L1 intent and L2 typed DAG commit to a topology before any codegen. The L4 deployment brief is structured: it names which artifacts to produce, which skill fields to cite, and which acceptance checks must pass. The agent edits within these contracts; an architectural pivot is structurally impossible without a pattern-layer (L2) revision, which is owned by the user, not the L4 sub-agent. T2 fixes stay at the L4 codegen level (e.g., a smoke-query fix), not architectural.

\subsection*{Failure trajectory 3: same error, five iterations (run 7)}

\noindent
\textbf{Outcome:} 5 iterations, 326 turns, $\sim$35 min wall; T0 \checkmark, T1 \ding{55}, T2 \ding{55}.

\begin{itemize}[leftmargin=*]
\itemsep0pt
\item \textbf{Iter 1:} 64 turns, 0 in-scope edits (T0 \ding{55}). Initial output incomplete.
\item \textbf{Iter 2:} 131 turns, 20 in-scope edits. T0 \checkmark, T1 \ding{55} with \texttt{redpanda is unhealthy}.
\item \textbf{Iter 3--5:} 34 + 44 + 53 turns; 1 + 1 + 1 in-scope edits per iteration. The same \texttt{redpanda is unhealthy} error recurs every iteration. The agent identifies the symptom but cannot find the root cause from the failure feedback alone.
\end{itemize}

\noindent
\textbf{Root cause.} The symptom is observable (Redpanda unhealthy) but the cause (a specific incompatible broker flag, or a missing network exposure) is invisible to the agent unless it reads the container logs directly. The harness's failure feedback strings the docker compose stderr, which only reports the symptom. Iteration in the absence of root-cause attribution becomes a guessing loop.

\noindent
\textbf{How DDS prevents this.} The L4 attribution loop \emph{types} the runtime signal. A persistent \texttt{healthcheck\_unhealthy} on a known system is classified as \texttt{L3\_skill} (composition gap) rather than \texttt{L4} (codegen slip), and the framework routes it to a skill-field edit instead of a codegen retry. Either the skill already covers the flag (the loop closes) or the skill is patched once (and every future deployment of that system inherits the fix; \S\ref{subsec:attribution-and-learning}).

\subsection*{Success trajectory: convergence in 2 iterations (run 10)}

\noindent
\textbf{Outcome:} 2 iterations, 77 turns, $\sim$16 min wall; T0 \checkmark, T1 \checkmark, T2 \checkmark; \texttt{stop\_reason=passed}.

\begin{itemize}[leftmargin=*]
\itemsep0pt
\item \textbf{Iter 1:} 27 turns, 17 in-scope edits. T0 \checkmark, T1 \checkmark (the stack actually boots), T2 \ding{55} with \texttt{smoke\_query\_error}: the verifier query references a column the agent named slightly differently in the DDL.
\item \textbf{Iter 2:} 50 turns, 6 in-scope edits. T0 \checkmark, T1 \checkmark, T2 \checkmark. Agent reconciles column names across the producer, DDL, and verifier.
\end{itemize}

\noindent
\textbf{Why it works.} This is the canonical pattern where iteration helps: iter 1 produced a bootable stack, the failure was a localized name mismatch, and iter 2 made the focused fix without restructuring anything. No architectural pivot, no symptom-vs-root-cause gap.

\noindent
\textbf{Comparison with DDS.} The same workload took DDS a median of 17 turns at \$1.49 to reach T1, with separate runs reaching T2; the iterated raw agent here used 77 turns at $\sim$\$3 to reach T2 on this single successful run. The other A success (run 6) needed 3 iterations and 289 turns. The framework contribution is not that iteration cannot work (it can, sometimes), but that the per-system composition knowledge an iterated agent must rediscover at every run is, in DDS, a reusable artifact (the L3 skill) edited once and cited inline thereafter.

\subsection*{What this tells us}

Three observations compress the A trajectories. (i) The dominant failure mode under iteration is not the absence of feedback but the inability to act on it: when the symptom is generic (\texttt{X is unhealthy}) and the root cause is in the system's own log, iteration loops without progress (run 7). (ii) Iteration can actively regress a stack when the agent's response to a downstream failure is an architectural pivot rather than a focused fix (run 5); a contract that pins L2 prevents this. (iii) When iteration does converge, it does so on a localized error class (smoke-query column mismatch, run 10) at substantial cost in turns; that cost is exactly what DDS's skill artifact amortizes across deployments.

\section{Supplementary figures from the trading case study}
\label{app:supplementary-figs}

This section collects supplementary material referenced from the main body. Figure~\ref{fig:intent-yaml} shows the typed intent for the trading workload (\S\ref{sec:case-study}); the headline pass rates and the attribution-and-learning loop are reported in tabular form (Tables~\ref{tab:headline} and~\ref{tab:failures}).

\begin{figure}[h]
\centering
\begin{minipage}{0.92\linewidth}
\small
\begin{verbatim}
intent:
  data_model:
    entities: [market_tick, ohlcv_bar, position, order]
    primary_types: [time_series, relational, event]

  access_pattern:
    read:  [olap_range_scan, point_lookup, streaming]
    write: [high_throughput_append, transactional_update]

  scale:
    ingest_rate_events_per_sec: 100
    retention_history_years: 5
    concurrent_users: 1

  latency:
    point_lookup_p99_ms: 10
    analytical_query_p99_ms: 2000

  consistency:
    ohlcv_aggregate: eventual
    positions: strong

  cost:
    monthly_usd_budget: 100       # ceiling: build + maintain
    preference: simplicity        # soft warning: under-specified
\end{verbatim}
\end{minipage}
\caption{Trading-workload intent populating all six L1 dimensions (\S\ref{subsec:intent}). The framework emits one soft warning (cost preference under-specified, defaulted to \emph{simplicity}) and no hard errors.}
\label{fig:intent-yaml}
\end{figure}

\section{DDS-Bench: Scope and Second-Domain Replication}
\label{app:bench-scope}

The trading case study (\S\ref{sec:case-study}) is one intent in DDS-Bench, the public benchmark specification we built to score any conforming framework that takes a typed intent and produces a multi-system backend. The bench is a specification rather than a runtime: it contains the typed inputs (intents, topology constraints, acceptance criteria), the workloads and probes a runner exercises, the metric vocabulary against which probes are scored, calibration material (golden traces at a fixed seed), and a \texttt{SystemAdapter} contract any framework conforms to (DDS, a raw coding agent, an IaC backend, a vendor stack). The current scope is 6 intents across 5 workload domains (\textbf{trading}, \textbf{ecommerce}, \textbf{chat}, \textbf{rag}, \textbf{iot}; \texttt{trading-binance} is the case-study intent), 30 declarative workloads with 30 byte-identical golden traces at \texttt{seed=42}, 59 probe templates that exercise the standard composition-stress dimensions (smoke, load, correctness, freshness, resilience), 21 reusable metrics (latency, throughput, no-loss, exactly-once, recovery, recall@k, cache-hit, ordering, encryption, oversell, downsample, etc.), and 8 JSON schemas for non-Python runners. The intent is to stress framework behavior along axes the trading workload alone does not hit: ecommerce stresses transactional consistency under burst writes, chat stresses ephemeral-vs-persistent boundaries and operator-algebra extensibility (\S\ref{app:chat}), rag stresses approximate-search recall and embedding-pipeline composition, and iot stresses high-fanout ingest at constrained bandwidth.

DDS-Bench is also the substrate on which a second-domain replication of the headline result becomes possible without rewriting the framework. We re-ran the same DDS pipeline (same skill catalog, same intent-to-deployment contracts, no L1--L4 changes) on the \texttt{ecommerce-mvp} intent: across $n{=}5$ runs, every run reached T0, T1, T2, and T3 (the SLO acceptance tier the trading case study defers to L5 for the trading intent specifically), with p99 latencies clustered tightly in the 1.4--2.7~ms range across 25 probe measurements; this is consistent with the trading-intent T0--T3 result and rules out one form of single-domain artifact (that the framework only handles trading-shaped topologies). What the second-domain run does \emph{not} settle is the open-set generalization questions named in \S\ref{sec:discussion} and Appendix~\ref{app:open-problems}: it is the same model, the same operator algebra, and skills authored against domains close to ecommerce. Cross-model replication (Condition C on a non-Anthropic model), broader-domain replication (chat, iot, rag at quantitative scope), and the open question of out-of-class L4 attribution remain in flight and are the priority items DDS-Bench is built to support.

\section{Extended Discussion: L5 Evaluator Layer}
\label{app:l5}

This appendix expands the L5 evaluator roadmap sketched in \S\ref{sec:discussion}. L5 is a design proposal, not a layer implemented in the prototype evaluated in \S\ref{sec:case-study}; the descriptions below use the conditional or future tense to make this explicit.

\paragraph{What the framework would own.}
A per-intent brief would derive mandatory probe dimensions from the intent: every declared access pattern would have a query path, every declared SLO on latency, throughput, no-loss, and freshness would be measured, and declared capacity would be compared against measured footprint. The framework would also own an access contract for what the evaluator may read or mutate, a report schema, and benchmark-failure-to-layer attribution rules that would route a probe failure to the layer owning the violated decision, parallel to the L4 loop.

\paragraph{What the agent would own.}
The evaluator sub-agent would generate per-domain artifacts (load driver, correctness probe, resilience probe), each citing the skill fields it targets, and a small hand-authored golden probe set per domain would serve as cross-validation, catching agent regressions on the probes themselves. The evaluator would mix three modes that each catch a distinct failure class: black-box (produce traffic at declared ingress, observe at declared egress, assert against the user contract), white-box (read intent, plan, deployment, and skills to check realizability before any traffic), and gray-box (container logs, metrics endpoints, and system tables to cross-check declared against actual). The agent's contribution over a fixed harness is that it would compose intent-specific semantic probes alongside conventional metrics: OHLCV bar continuity and aggregate correctness in trading, message-ordering preservation under retransmission in chat, and similar domain-meaningful invariants that no design-time benchmark suite ships.

\paragraph{Why L5 leads the roadmap.}
L5 would close the architecture: the L4 loop today carries only crash-time signals, while the evaluator would surface SLO-time and correctness-time signals that route back to every earlier layer (capacity-vs-footprint mismatch to intent revision at L1, SLO violation to plan alternatives at L2, anti-pattern hits to skill patches at L3, codegen slips to code patches at L4), and T3 (declared SLOs holding under load) would become the first acceptance tier the evaluator makes routine.

\paragraph{Production-grade gaps and where they land in DDS.}
Where we are: the prototype runs on local Docker Compose and clears a smoke-level acceptance bar (boot, healthchecks, end-to-end query) on a single intent in a single host environment; it is proof-of-life, not production-grade. The gap to a real backend spans five concern groups: behavior under load (load testing at declared scale, backpressure, cost measurement under real workload), correctness under failure (durability, replay semantics, delivery guarantees, persistent-volume recovery), operational lifecycle (schema migration, high availability), broader deployment scope (cloud targets beyond Compose, secrets, security boundaries), and observability (alerting, SLO cross-checking). Each of these lands at a specific DDS layer rather than at the architecture itself. Load and observability concerns are L5 evaluator targets, with signals routing back to L1 (intent revision) and L3 (skill patches). Failure-correctness and lifecycle concerns express as L1 declared dimensions plus L3 skill content. Cloud deployment beyond Compose is an L4 codegen target rather than an architectural gap, since the L3 plan is product-list-plus-config and neutral to the runner. Security and secrets are a cross-cutting contract over L1, L3, and host policy, which the chat case study (Appendix~\ref{app:chat}) flagged as architectural rather than as a checkbox. Closing these gaps is engineering inside the existing typed contracts, not restructuring of those contracts.

\paragraph{Other roadmap axes.}
At L2, the operator algebra is open: each new application domain adds operators (the chat platform in Appendix~\ref{app:chat} added \texttt{ROUTE}, \texttt{NOTIFY}, and \texttt{INDEX}), and a principled path from these additions back into the L2 type system, with composition-rule inference rather than hand edits, is the next algebraic step. A cost-objective formulation over multi-system topologies, analogous to a physical query optimizer but across products rather than within one, is a natural extension that takes its cost data from L3 skills and its ranking from the L2 planner. Skills are expert-authored today, and the next phase is to extract composition rules and anti-patterns semi-automatically from documentation, incident reports, and post-mortems, with the combined L4 attribution log and L5 evaluator reports as training signal. Two further extensions add constraints rather than capabilities and stay within the existing structure: multi-tenant deployment and cross-cloud data residency contribute intent dimensions at L1 and composition constraints at L3, and security and governance, which the chat case study flagged as architectural rather than as a checkbox, become a contract layer cross-cutting L1 and L3 over who may hold what key and where data may live.

\section{Extended Open Problems by Layer}
\label{app:open-problems}

This appendix expands the open problems sketched in \S\ref{sec:discussion}. Two questions live at L1 because the L1 contract has both a static intent specification (subject to drift) and a draft-producing elicitation sub-step (depicted as L0 in Fig.~\ref{fig:layers}); the others sit at L4 and L3.

\paragraph{L4: attribution confidence and out-of-class behavior.}
The $n{=}20$ in-distribution check in \S\ref{subsec:attribution-and-learning} confirms only that the rule set fires on the five classes it was authored against (a sanity check, not a generalization claim). Out-of-class behavior is the central open question: when the classifier sees a fault outside the five enumerated classes (e.g., TLS handshake failure, vendor rate-limit, clock-skew), the framework's safe default is to fall through to a generic ``acceptance-failure'' label that re-routes to a focused L4 edit with the raw signal attached, rather than to a confident wrong attribution. The acceptable cost of that fall-through, and the false-positive rate of the in-class rules on adjacent signals, are not yet measured. Two further sources of ambiguity sit alongside the open-set problem. First, some runtime signals (consumer-lag, p99 violation) are genuinely ambiguous between L2 and L3: they may attribute to a pattern choice (the topology cannot meet the SLO) or a product choice (this product cannot meet the SLO under this configuration); the harness marks these ambiguous and accepts either label, but a principled confidence model that decides when to apply a skill patch automatically, when to ask the user, and when to consider a re-plan remains to be designed. Second, signals correlated across multiple containers (a cascade) are routed by the strongest single signal today, with no aggregation across container boundaries. The natural training signal for both is the attribution log itself, paired with the eventual fix and the post-fix outcome; the L5 evaluator above is the natural source of richer labeled signals once it lands.

\paragraph{L1: intent drift.}
A real workload evolves: traffic mix changes, consistency requirements relax or tighten, and an intent signed off today may mis-describe tomorrow's traffic. The right refresh cadence and the right owner of that refresh (user, planner sub-agent, or framework) is unsettled. Periodic re-validation against observed metrics is one direction; user-initiated revision is another; an evaluator-triggered drift signal (declared capacity vs.\ measured footprint) is a third. The cost of staleness is borne by every downstream layer, so the dial belongs above L1 rather than inside any sub-agent.

\paragraph{L3: skill versioning and deprecation.}
The systems beneath skills change underneath us: Kafka 3 to 4, ClickHouse storage formats, vendor image deprecation, client-library breaking changes. Skill fields age at very different rates, and the four-block structure (\S\ref{subsec:skills}) lines up with the rates we observed across the trading and second-domain runs. The fastest-aging fields are pinned image tags in \texttt{operational.recommended\_images} (Kafka and ClickHouse each pushed a minor release inside our case-study window), followed by host-environment specifics in \texttt{operational.known\_host\_port\_conflicts} (a different host shifted the F2 conflict to a new port; \S\ref{subsec:attribution-and-learning}); anti-pattern severity in \texttt{anti\_patterns} ages next, as a hard limit can soften when an upstream fixes the underlying issue (the TTL-on-DateTime64 entry in Appendix~\ref{app:skill-yaml} is exactly such a candidate as ClickHouse evolves); \texttt{capabilities} ages slowest with the system itself. A per-deployment freeze of the catalog state (currently a content-hashed lock file alongside the skill directory) gives us reproducibility per run, but a principled versioning and deprecation policy that lets old skills age out across $N$ deployments without losing the attribution-log history attached to them is open. The naive solution (rewrite the skill) loses the trail of which past failures motivated which fields. A better approach versions fields rather than files and preserves the lineage from each runtime signal to the field it patched. Measuring field-staleness rates is a natural use of the attribution log itself: a field never re-cited after $k$ deployments is a candidate for promotion to \texttt{capabilities}, and a field re-patched at a high rate is a candidate for splitting into a finer-grained sub-field.

\paragraph{L1 (elicitation sub-step): protocol.}
A draft intent is only as good as the dialogue that produced it. The trade-off between dialogue length and intent coverage across the six L1 dimensions has not been measured. The right protocol minimizes user effort while maximally covering the six dimensions and surfaces under-specification cheaply (the cost-preference soft warning in the trading walkthrough is one example). How to bound the dialogue without leaving dimensions blank, how to handle conflicts between user-stated and inferred values, and how to elicit numeric envelopes without forcing premature precision are open design questions.

\section{Extended Related Work}
\label{app:extended-related}

\paragraph{Agentic discovery systems.}
DDS sits in the recent line of agentic-discovery systems that impose structure on an LLM-driven search space. AlphaEvolve~\citep{novikov2025alphaevolve} structures evolutionary search over programs with benchmark verifiers; SkyDiscover~\citep{liu2026skydiscover}, OpenEvolve~\citep{sharma2025openevolve}, and ShinkaEvolve~\citep{lange2025shinkaevolve} extend the same paradigm to broader algorithmic and scientific discovery. GEPA~\citep{agrawal2026gepa} structures prompt-optimization with reflective natural-language attribution from agent traces; ACE~\citep{zhang2026ace} structures context evolution; DSPy~\citep{khattab2024dspy} compiles declarative LM-module graphs; Meta-Harness~\citep{lee2026metaharness} jointly optimizes the harness. Glia~\citep{hamadanian2025glia} is the closest sibling on the systems side, structuring multi-agent reasoning over distributed-systems designs to produce expert-level configurations. DDS extends this paradigm to a different search space (multi-system data backends) with a different verifier (deployment outcome rather than a benchmark) and a different memory unit (per-system editable skill artifacts instead of per-task prompts). Empirical work on agent reliability (multi-agent failure modes~\citep{cemri2025mast}, multi-agent under-performance~\citep{pappu2026multiagentteams}, production-agent reliability~\citep{pan2025measuringagents}) grounds the architectural target: typed failure routing across system boundaries and composition knowledge as an editable, citable artifact.

The systems below are \emph{adjacent} to DDS rather than closest neighbors: most operate at a different abstraction level (operators inside one engine, contracts inside one warehouse, knobs inside one product, queries against one assumed schema, code edits inside one repository) than DDS's cross-system composition target. Coding agents like Claude Code (\S\ref{sec:motivation}) are the closest neighbor on the deployment side; Glia (above) is the closest on the agentic-discovery side. Everything below sits one level over from those.

\paragraph{LLM-native pipelines (different abstraction level).}
LOTUS~\citep{patel2025lotus}, DocETL~\citep{shankar2025docetl}, and Palimpzest~\citep{liu2025palimpzest} introduce declarative semantic operators (map, filter, join, aggregate over unstructured data) with cost--quality optimization inside a single LLM-native engine. These are declarative over \emph{operators that run LLMs on data}, while DDS is declarative over \emph{which heterogeneous systems implement which operators across a multi-system backend}. The two abstractions compose, and a DDS \texttt{TRANSFORM} node can be backed by such a semantic-operator pipeline at L3, but they sit one level apart and are not the same problem.

\paragraph{Cross-project composition in the modern data stack (extended).}
dbt Mesh~\citep{dbtmesh} adds typed contracts and cross-project references on top of dbt, lifting the dbt model from a single project to a federation of projects with explicit producer/consumer contracts; Apache Iceberg~\citep{iceberg} provides a typed table format with schema and partitioning evolution that several engines can share. Both target the boundary problem DDS occupies but stay within the data-warehouse perimeter: contracts are expressed between SQL projects or between query engines reading the same table format, not between heterogeneous systems (a queue, an OLTP store, a cache, a search index) chosen against an intent. DDS's L2 operator DAG and L3 skill contract are designed exactly to cross those system boundaries, and an Iceberg table or a dbt-Mesh contract is a natural physical-layer instance at a DDS \texttt{STORE} or \texttt{TRANSFORM} node.

\paragraph{Coding-agent landscape (extended).}
Beyond the benchmarks anchored in \S\ref{sec:motivation}, the coding-agent landscape also includes SWE-agent~\citep{yang2024sweagent} (agent-computer interface scaffolding), MLE-bench~\citep{chan2024mlebench} (ML engineering tasks), and DS-1000~\citep{lai2023ds1000} (data-science code). All measure competence inside a single repository, notebook, or task; none tests whether a declared multi-system backend can boot and stay healthy end-to-end under typed failure attribution, which is what our T1 evaluates.

\paragraph{Federated query and HTAP engines.}
Federated query engines such as Presto/Trino~\citep{sethi2019presto} and Spark~SQL~\citep{armbrust2015sparksql} extend the polystore idea to execution over pre-existing stores, and HTAP / ``NewSQL'' systems (HyPer~\citep{kemper2011hyper}, SAP~HANA~\citep{farber2012hana}, Google~F1~\citep{shute2013f1} over Spanner~\citep{corbett2012spanner}) consolidate operational and analytical workloads within one product. In DDS these engines appear as candidate L3 products at individual operators; the framework chooses among them under a typed intent and skill contracts rather than assuming a single target.

\paragraph{Self-driving databases beyond one-knob tuning.}
Peloton~\citep{pavlo2017peloton} and CDBTune~\citep{zhang2019cdbtune} extend the self-tuning line beyond OtterTune's config search to physical design and deep-RL knob policies, still within one product. DDS differs in two ways: the attribution loop produces \emph{cross-system} patches (e.g., a Kafka-side retention change driven by a ClickHouse-side signal), and the unit of learning is an editable agent skill reviewed like code rather than a black-box policy.

\paragraph{Text-to-SQL and declarative foundations.}
Text-to-SQL benchmarks and systems such as Spider~\citep{yu2018spider}, BIRD~\citep{li2023bird}, and DIN-SQL~\citep{pourreza2023dinsql} translate natural-language queries over an \emph{assumed} schema. DDS addresses the upstream problem of composing a schema and backend that can serve the intent in the first place, and a text-to-SQL pipeline could itself be the product at a \texttt{SERVE} node. The declarative-what, imperative-how separation~\citep{codd1970,chamberlin1974sequel} is the conceptual ancestor of L1 and L2, which DDS lifts from within-product queries to across-product composition.

\paragraph{AI-driven systems research and accountability.}
Inefficiencies of Meta Agents~\citep{el2025metaagents} argues against fully-automated meta-agent design loops on cost and behavioral-diversity grounds, supporting our position that human-authored declarative skill artifacts are the right unit of composition knowledge. AI-Driven Research for Systems (ADRS), introduced by ``Barbarians at the Gate''~\citep{cheng2025adrs} and extended by ``Let the Barbarians In''~\citep{cheng2025letbarbariansinai} to systems performance research, argues AI is upending systems research methodology by exploiting cheap reliable verifiers; DDS's T0/T1/T2 acceptance gates are an instance of the same pattern at composition time, where the verifier is a runnable backend rather than a benchmark target. Cost-of-Pass~\citep{erol2025costofpass} formalizes accuracy--cost tradeoffs in LM evaluation; our cost-and-turn measurements in Table~\ref{tab:headline} adopt the same accountability stance.

\paragraph{Persistent, editable skill libraries for LM agents (foundations).}
The idea of replacing ephemeral in-context learning with a persistent, editable skill library has roots that predate DDS. Voyager~\citep{wang2023voyager} introduces a lifelong-learning embodied agent in Minecraft whose skill library accumulates discovered behaviors as reusable code that the agent re-uses on later tasks. Subsequent work on \emph{agent skills}~\citep{anthropic2025agentskills} casts the same idea as an OS-level facility: a structured, versioned bundle that an agent loads on demand and edits over time. DDS's L3 agent-skill artifact (\S\ref{subsec:skills}) is in this lineage; the contribution is not the editable-library idea itself but its application as the unit of \emph{composition knowledge} for multi-system data backends, with the framework shape (typed L1--L4 contracts and L4 attribution) providing the channels by which skill edits enter and survive across deployments.

\paragraph{Multi-agent IaC generation (MACOG and the IaC-with-LLM line).}
MACOG~\citep{khan2025macog} is structurally the closest neighbor on the multi-agent-with-deploy-feedback axis: it decomposes Terraform generation into eight specialized roles (architect, provider harmonizer, engineer, reviewer, security prover, cost/capacity planner, DevOps, memory curator) coordinated on a shared blackboard, with Terraform Plan and Open Policy Agent as the verification signals, and reports IaC-Eval~\citep{kon2024iaceval} gains over a RAG baseline. The structural building blocks overlap (typed contracts between agents, a Memory Curator as the persistence unit, deploy-time feedback as the convergence signal), and the comparison is informative because it isolates what the layer-of-abstraction choice changes once those building blocks are taken as given. MACOG operates \emph{inside one IaC dialect}: every role generates or critiques Terraform for a chosen cloud, and the deploy-feedback signal is Terraform Plan over that already-chosen architecture. DDS operates \emph{above} that level: the operative choice is which heterogeneous data systems (Kafka vs.\ Redpanda vs.\ NATS at QUEUE; ClickHouse vs.\ TimescaleDB vs.\ DuckDB at STORE-analytics; an OLTP, a cache, an index, and so on) compose into a topology that meets a typed intent, and the verifier is whether that composition boots and meets declared SLOs end-to-end rather than whether a single IaC artifact validates. The two are stackable rather than competing: a MACOG-class agent is a natural realization of a DDS \texttt{L4} sub-agent for a chosen architecture (the L3 plan as a product-list-plus-config feeds an IaC backend), and conversely a DDS-class framework is a natural composition layer over MACOG-class IaC generators. The orthogonal contribution claims also separate: MACOG's gain is over a within-Terraform RAG baseline on a benchmark of single IaC artifacts; DDS's gain (\S\ref{subsec:headline}) is over iterated raw-agent baselines on whether a multi-system stack boots and serves end-to-end, where the failure modes the structure prevents (an architectural pivot under a downstream signal, an iteration loop on a system-internal log) do not arise inside a single IaC dialect.

\end{document}